\documentclass{article}
\usepackage{graphicx} 
\usepackage{style}
\usepackage{url}
\usepackage{hyperref}
\usepackage[utf8]{inputenc}
\usepackage{float}
\usepackage{verbatim}
\usepackage{comment}
\usepackage{placeins}

\title{Data Fusion for Multi-Task Learning of Building Extraction and Height Estimation}
\author{saadahmedjamal}
\date{February 2023}
\name{Saad Ahmed Jamal, Arioluwa Aribisala}
\address{Université Bretagne Sud France, Paris Lodron Universität Salzburg Austria}
\begin{document}

\maketitle

\begin{abstract}
In accordance with the urban reconstruction problem proposed by the DFC'23 Track 2 Contest, this paper attempts a multitask-learning method of building extraction and height estimation using both optical and radar satellite imagery. Contrary to the initial goal of multitask learning which could potentially give a superior solution by reusing features and forming implicit constraints between multiple tasks, this paper reports the individual implementation of the building extraction and height estimation 
under constraints. The baseline results for the building extraction and the height estimation significantly increased after designed experiments.
\end{abstract}
\keywords{data-fusion, segmentation, dsm, regression}
\section{Introduction}

The study of buildings is central to urban development and restructuring. With increasing research in the deep learning domain and an increase in state-of-the-art deep learning methods, there has been a focus on the automatic extraction of building footprints and classification of buildings using either aerial or multi-spectral satellite images. A challenge that remains largely unsolved is the estimation of the heights of the extracted building in imagery. This information is however very important for building reconstruction through the analysis of satellite imagery and deep learning techniques. A number of works have been done in the area of height estimation using stereo images, and single-view aerial imagery but most research works do not address the use of multi-view. 
The 2023 IEEE GRSS Data Fusion Contest (DFC23) is a competition organized by a group of organizations to encourage more research in building extraction, classification, and 3D modelling. The goal is to create more accurate models of urban areas that include information about different types of roofs. The competition created a comprehensive benchmark for the classification of building roofs that includes many features. It includes two competition tracks that use both optical and Synthetic Aperture Radar (SAR) data to focus on roof type classification and building height estimation. The data provided is unique because it is a large-scale, globally distributed urban building dataset that covers seventeen cities across six continents, with a wide range of building styles. The dataset includes nearly 300k instances of buildings with fine-grained labels for twelve different types of building roofs. The use of both optical and SAR images allows for multimodal data fusion that can lead to more accurate models for building extraction and classification \cite{9857458}.

The height estimation baseline makes use of Autoencoders. Autoencoders are a type of neural network that is used for unsupervised learning, which means that they don't require labelled data to train. They are composed of an encoder network and a decoder network that work together to learn a compressed representation of the input data.
The encoder network takes the input data and maps it to a lower-dimensional latent space representation, also known as a bottleneck layer. This compressed representation should capture the most important features of the input data while reducing the dimensionality of the data.
The decoder network then takes this compressed representation and tries to reconstruct the original input data from it. The goal of the autoencoder is to minimize the difference between the original input and the reconstructed output. This is typically done using a loss function such as mean squared error.

PSPNet (Pyramid Scene Parsing Network) is a deep neural network architecture for the semantic segmentation of images. It was proposed in 2017 \cite{Zhao_2017_CVPR}. PSPNet and ASPNet both architectures were tried for the decoding part.
The PSPNet architecture also includes skip connections that allow low-level features to be combined with high-level features in the network. This helps preserve spatial information and improves the accuracy of the segmentation masks.
ASPNet (Attentional Spatial Pyramid Pooling Network) uses a novel attentional spatial pyramid pooling module that selectively emphasizes informative regions in the feature maps \cite{bruno2021combining}. This module helps to capture long-range dependencies and improve the accuracy of the segmentation masks.
Overall, both architectures have their strengths and weaknesses, and the choice between them depends on the specific requirements of the task at hand. PSPNet is well-suited for capturing global context and has achieved state-of-the-art performance on several benchmark datasets, while ASPNet is well-suited for capturing long-range dependencies and has shown promising results on various datasets. Both of them were implemented in the baseline. Which architecture is better suited for a particular task depends on many factors. For this task PSPNet gave better results than ASPNet, therefore it was used for later experiments.

\FloatBarrier

\section{Dataset}
The DFC'23 dataset from the high-resolution commercial Chinese satellites known as Superview-1 and Gaofen-2 by Beijing Space View Tech Co Ltd \cite{mrnt-8w27-22}. It contains panchromatic and multispectral images at a resolution of 0.5m and 0.8m pan-sharpened. It also contains Gao-fen-3 SAR data at 1m resolution. The normalised DSM provided as height labels were captured from stereo images by Gaofen-7 and WorldView 1 and 2 at a ground sampling distance of 2m resolution \cite{rs12172719}.

17 cities from 6 continents were sampled in the original paper, however, the track 2 data had only 11  countries represented. These cities include Berlin, Barcelona, Rio, Sydney, Brasilia, Copenhagen, New York, New Delhi, San Diego, Sao Luis, and Portsmouth. During the course of the training, the data was eventually beefed up to include track 1 data in an attempt to increase the accuracy of the model performance. Therefore, more cities were eventually represented such as Tokyo, Addis Ababa, Jacksonville, Sao Paulo, Darwin and Suzhou.

\begin{figure}
    \centering
    \includegraphics[width=3.3in]{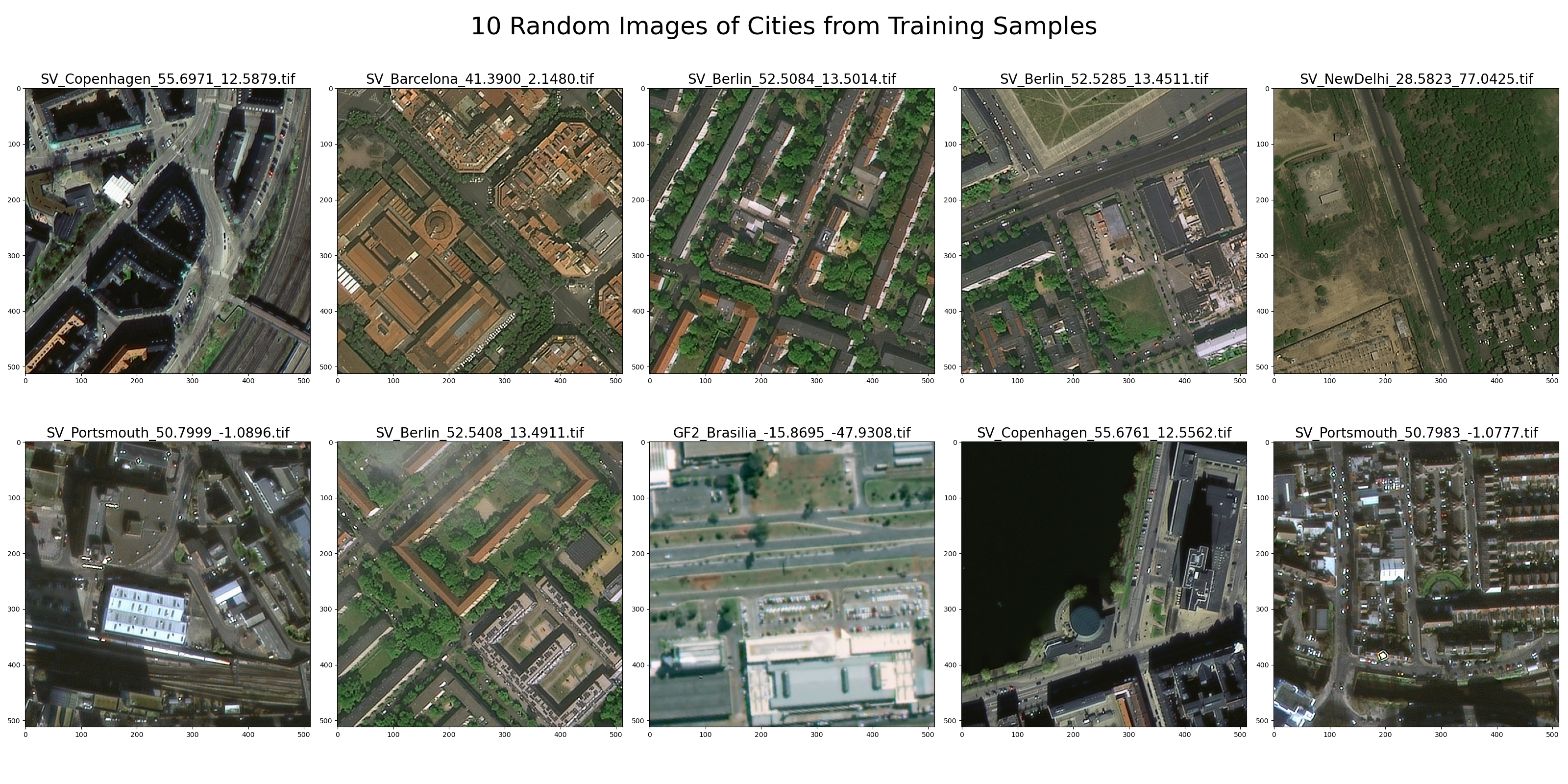}
    \caption{Training batch images of different cities}
    \label{fig1}
\end{figure}

\begin{figure}
    \centering
    \includegraphics[width=3.3in]{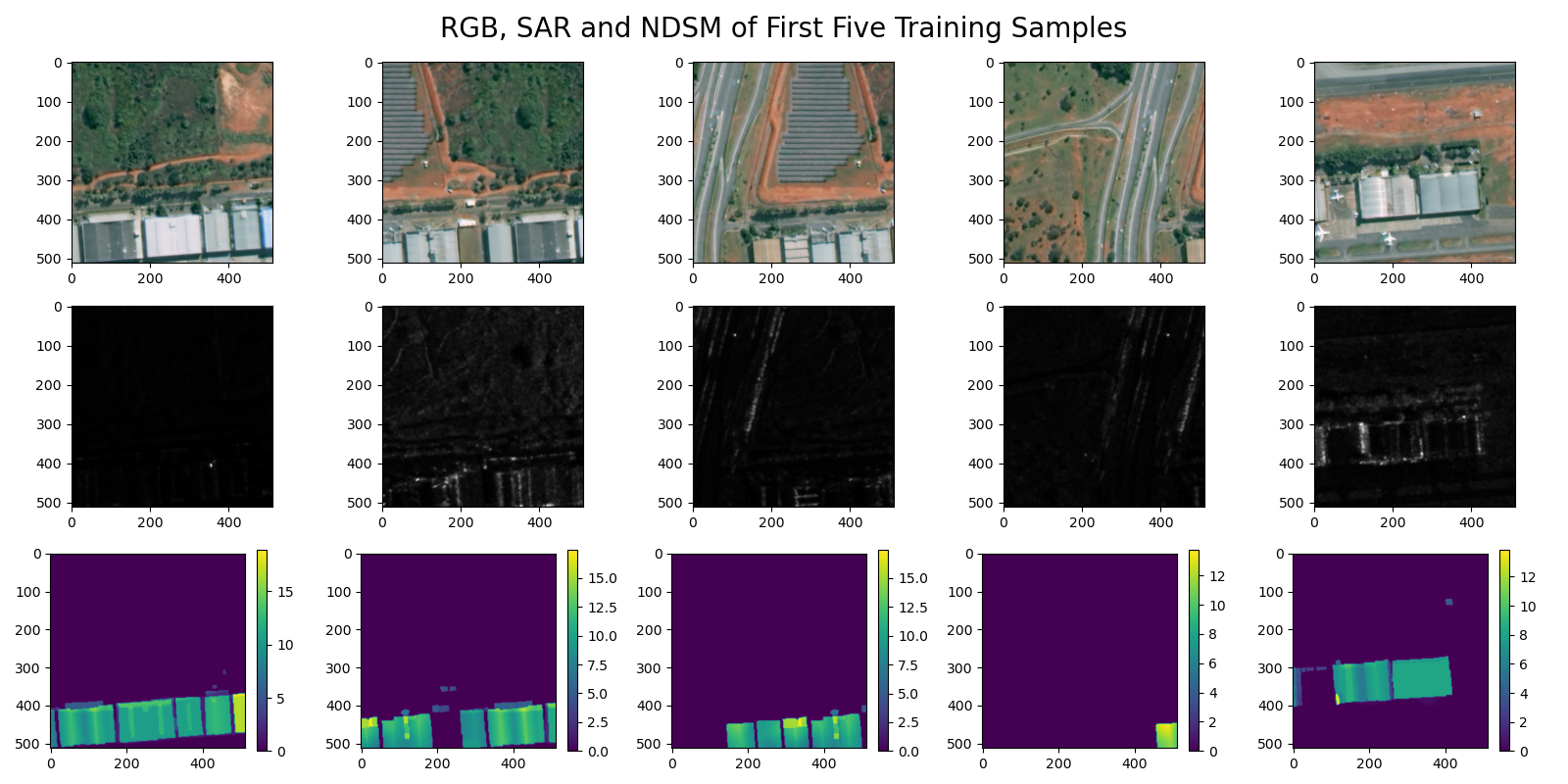}
    \caption{Training Samples}
    \label{fig2}
\end{figure}

\section{Methodology}

The experiments were done using the conventional method of running the building extraction and height estimation separately. This was because of the implicit challenge of predicting the height of the whole image and not just the building as seen in the normalised DSM provided as labels for the height estimation task. The baseline scripts were given for each task with different deep learning architectures for each task. The building extraction baseline is based on the classical mask-rcnn algorithm within a robust image segmentation and object detection toolbox known as mmdetection by the open-mmlab \cite{He_2017_ICCV}. 
On the other  hand, the building height estimation baseline was not dependent on any robust customized toolbox. Each task had baseline results as seen in table one that needed to be improved on.
\begin{table}
\caption{\label{table1} Baseline Results }
\begin{center}
\begin{tabular}{ |c c | c c| } 
\hline
  Type & Modality & mAP & mAP50 \\
 \hline
 Mask R-CNN & RGB & 22.8 & 49.1 \\ 
 Mask R-CNN & RGB+SAR & 18.6 & 43.6 \\ 
 \hline
\end{tabular}
\end{center}
\end{table}
In order to improve the results of both tasks, first we carried out a sensitivity analysis. This involved experimenting with several hyperparameters to choose the optimal hyperparameters. Some of the hyper-parameters that were tested were the batch size, the loss function, the optimizer (such as Adams and Stochastic Gradient Descent), the number of epochs and the learning rate.  
Thereafter, different resnet variants (resnet-50 and resnet-101) were experimented with to select the best-performing backbone architecture. The pre-trained weights of these models were used. 
For the building extraction task, several segmentation libraries were experimented on such as                           Mask-R-CNN, Cascade Mask-RCNN, Mask2former and Retinanet. The Mask-R-CNN classical method was however settled for because it gave a stable prediction and was able to improve the results better than the others.

Thirdly, experiments were conducted using separate modalities to compare results. These included the comparison between the results of using only RGB images or only SAR images. Using RGB images as input gave better results.
Fourthly, data fusion techniques were applied for both the building extraction and the height estimation tasks.
\begin{enumerate}   
\item Early fusion 
\item Intermediate fusion
\item Late fusion
\end{enumerate}

Lastly, the introduction of skip connections to the network in the height estimation task was done to improve the predictive performance of the height estimation models.

1. Early Fusion: 
Early data fusion in deep learning refers to the process of combining different sources of data at an early stage of a deep neural network model, typically before the first layer of the network. This is in contrast to late fusion, which combines the output of multiple models at a later stage, such as after the last layer of a network. In early data fusion, the multiple sources of data can be of different modalities, such as image, text, and audio, and the goal is to create a unified representation of the input data that can be used for downstream tasks, such as classification or regression. This approach can be particularly useful when the different modalities contain complementary information that can help improve the accuracy of the model. In this experiment, RGB and SAR were stacked together before being fed into the model for training. This caused an increase in the dimensionality of the input data. This is due to the number of channels supported by the resnet pre-trained backbone architecture used as a result of its prior training on imagenet RGB images. To increase the modality from 3 dimensions to 4 dimensions to accommodate the concatenation of an additional SAR channel, the model was customized. 
This method of data fusion was experimented on both the height estimation task and the building extraction tasks. 

2. Intermediate Fusion:
In deep learning it refers to the process of combining features extracted from different modalities at an intermediate stage of a neural network model, typically after the first few layers. This is in contrast to early fusion, which combines the input data from different modalities before the first layer of the network, and late fusion, which combines the output of multiple models at a later stage.
Intermediate fusion is often used when the different modalities contain different levels of information, and it is not clear which modality should be given more weight in the early stages of the model. By combining the features extracted from different modalities at an intermediate stage, the model can take advantage of the complementary information contained in each modality while also learning which modality is more informative for the task at hand.
One common technique for intermediate fusion is to use a multi-modal CNN architecture with multiple branches, where each branch processes the input from a different modality and shares some layers with the other branches. The outputs of the shared layers are then concatenated and passed through additional layers to generate the final output.

3. Late Fusion:
In late fusion, the outputs of several models or features are combined using a fusion strategy such as averaging, concatenation, or weighted sum. 
For the late fusion experiment, the regression output of models were merged by taking average so an average of the two separate models were used.

\begin{figure}
    \centering
    \includegraphics[width=3.3in]{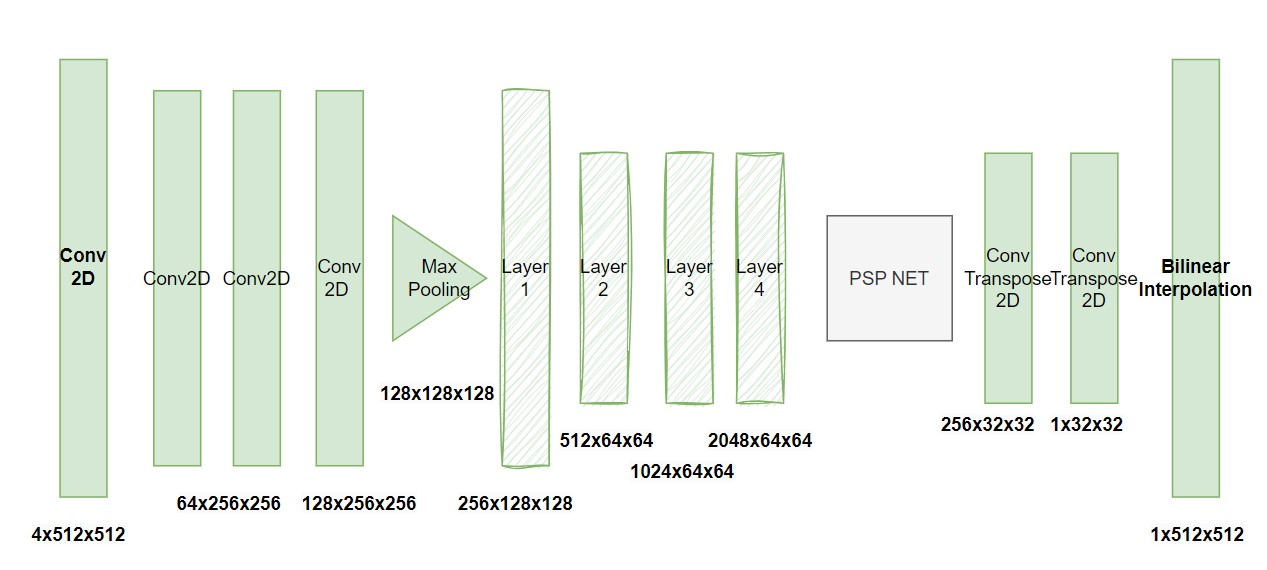}
    \caption{Model Architecture}
    \label{fig3}
\end{figure}

For the building height estimation, all three data fusion methods were experimented on. A few alternate decoding strategies were also studied. As shown in figure \ref{fig3}, decoding is mainly done through bi-linear interpolation. Model predicts 32x32 image size which is then expanded onto original 512x512. Attempts were made to involve transpose convolutions. Though, those strategies show good overall root mean square error, yet they were not able to predict buildings height correctly. Due to continuous nature of terrain, optimal decoding strategy is to up-sample through interpolation.

The Baseline makes use of Deep Lab V3 Architecture \cite{8806244} for the height estimation. This deep learning architecture consists of auto-encoder architecture with Resnet-50 as the backbone with a customized bottleneck and PSP net as a decoder. Previously, it was used for segmentation \cite{7913730}. With the customization of the last layer, it was made to give height values as output. 

Smooth L1 Loss was used as the criterion. The evaluation metric for the height estimation was the metric of height estimation is threshold accuracy 
\$1=Nt⁄Ntotal,  
where Ntotal is the total number of pixels and Nt is the number of pixels that meet max(y/\^y,\^y⁄y)<1.25, where y is the 
reference height and \^y is the predicted height. The final metric is the average of the two metrics, i.e., (AP50 + \$1)/2.

As initially stated, for the building extraction task, the \href{https://mmdetection.readthedocs.io/en/latest/get_started.html}
{mmdetection} toolbox by the  \href{https://github.com/open-mmlab/mmdetection}{open-mmlab} was used. This provided a variety of pre-trained models for training and predictive inferences. These pre-trained models are present in the model zoo within the toolbox/library. To use these models, configuration files needed to be tailored to suit each prediction task based on the unique differences between the models. Some of the models explored during this experiment were Retinanet, Mask2former, Cascade Mask R-CNN and Mask R-CNN. 

The Mask R-CNN model is a two-stage object detector which uses a Region Proposal Network (RPN) in its first stage and in parallel predicts the class, bounding box and a binary mask for each ROI. It is an extension of the Faster R-CNN model which outputs a class and a bounding box. Mask R-CNN outputs a segmentation mask as an additional output. This mask is very different from the class and bounding box outputs because it requires the extraction of a finer spatial layout of objects in the image.
The Cascade Mask R-CNN pre-trained model, differs from the Mask R-CNN in the number of branches in parallel. In the Mask R-CNN architecture, the segmentation branch is inserted in parallel to the detection branch while in Cascade Mask R-CNN, multiple detection branches are utilised in the architecture. 
The Retinanet Architecture is a one-stage detector that utilises focal loss as a loss function. This architecture was not explored extensively during the experiments and as such was not used eventually to make inferences. Likewise, the Mask2former architecture performed poorly in the validation and was not followed through for predictive inferences. The Mask2former architecture (Mask Attention Mask Transformer) architecture extracts localized features by constraining cross-attention within the predicted mask region.
The Mask-RCNN model was used for the building extraction experiment. The Mask R-CNN model is pre-trained on the coco dataset and evaluated using the  COCO metric AP50 (the lou threshold is 0.5), where the loU is evaluated based on the masks converted from the ground truth masks and the masks result. The Stochastic gradient descent optimizer was used throughout the experiment after confirming that the Adams optimizer yields a lower result. All models were initially trained for 36 epochs with a batch size of 16. After more experimentation, the Mask R\_CNN with a resnet101 backbone yielded better results against the resnet-50 backbone. Hence, to improve the model, the model was trained for 50 epochs. Normalisations were also customised to according to the coco dataset mean and standard deviations.

\section{Results}

Several changes were made in the baseline which improved the results. To evaluate the performance of the model, predictions were made on the given validation set (which acts as a test set since the labels of the validation set were withheld). These predictions were submitted on the IEEE GRSS Data Fusion  2023 Contest Track 2 portal on coda-lab. The height estimation result is a map with a resolution similar to the input image. The result of the model shows the estimated heights per pixel. This output was saved as a TIFF file with the identical name as the corresponding optical image. The performance on the validation set was reported back in form of a text file containing the r2 score, mean absolute error and AP\_50 values. 
On the hand, the Mask R-CNN output 2 JSON files containing inferences for both the bounding boxes and corresponding segmentation. The results were evaluated based on the Average Precision scores of the bounding boxes and the segmentation masks (mAP, mAP\_50, mAP\_75, mAP\_s, mAP\_m, mAP\_l based on the thresholds computed.  For the submissions, only the segmentation JSON file was required and the most important evaluation score used for the submissions is the mAP\_50.

For the first submission using the baseline, the scores recorded for the building extraction and height estimation were mAP\_50 = 49.3 and an accuracy score of 24.0 respectively. The result of the height estimation showed a significant visual similarity in the prediction. However, the results were not promising with respect to evaluation metrics. With the early fusion of data, the height estimation increased to 30,5 as compared to the late fusion score which reduced to 29.3.
Figure \ref{fig5} shows predicted DSM for training and test dataset imagery. 
\begin{figure}[H]
    \centering
    \includegraphics[width=3.1in]{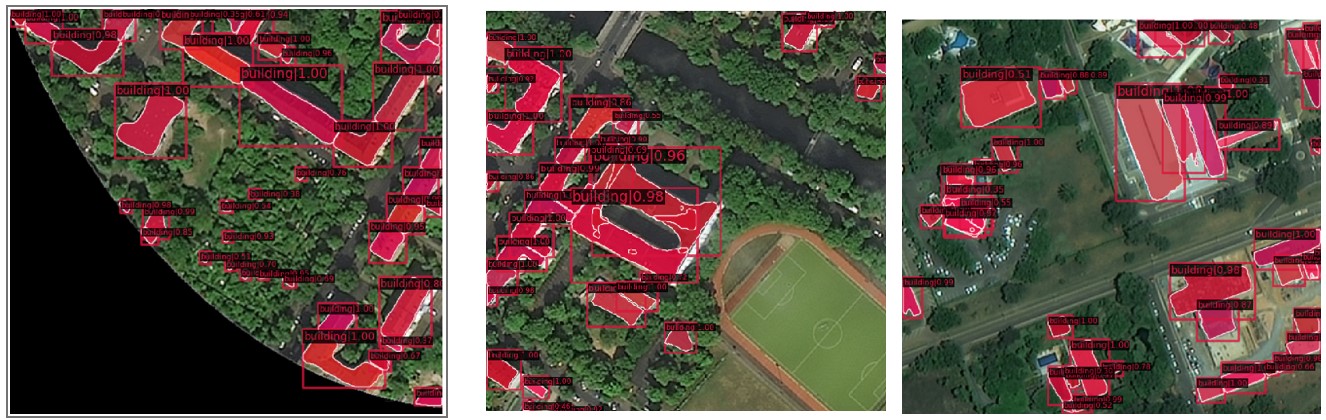}
    \caption{Building detection and segmentation }
    \label{fig4}
\end{figure}
\vspace{-5em}
\begin{figure}[H]
    \centering
    \includegraphics[width=3.3in]{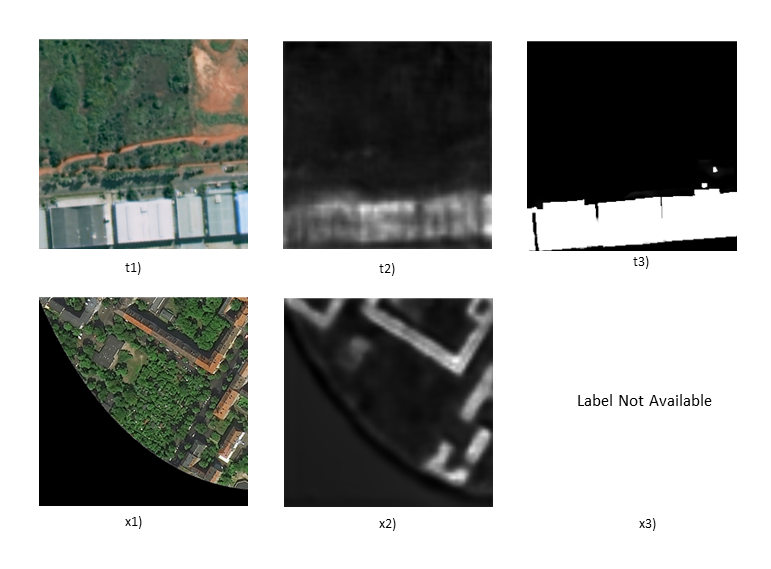}
    \caption{Height predictions for training and test t1) training RGB image, t2) prediction t3) label, x1) test RGB image, x2) prediction x3) label not available for test set }
    \label{fig5}
\end{figure}

After several experimentations, using different architectures such as Cascade Mask R-CNN, Mask2former and Mask R-CNN, the classical Mask R-CNN gave the best results comparatively. After experimentation, 
it was discovered that resnet101 backbone performed better than resnet50 backbone. Hence, Mask-RCNN using resnet101 improved the result of the RGB image input from mAP\_50 = 49.1 baseline result to mAP\_50 = 49.3 while in the early fusion of RGB + SAR the result improved from mAP\_50 = 43.6  to mAP\_50 = 49.6. 
To further improve the results. Data Augmentation techniques were employed, specifically by increasing the amount of training data. This was done by adding the data provided for track one (participants were not restricted to using only track2 data as confirmed in the forum). Increasing the data, therefore, increased improved the RGB result by 0.5\% while the RGB + SAR result improved by 0.4\%.

\begin{table}[H]
\begin{center}
\caption{\label{table2}Buildings footprints }
\begin{tabular}{ |c c | c| } 
 \hline
 Type & Modality & AP\_50 \\
 \hline
 Optical & RGB & 49.30 \\ 
 Optical (data augmentation) & RGB & 49.80 \\ 
 Optical cascade & RGB & 47.30 \\ 
 Early Fusion & RGB+SAR & 49.60 \\ 
 Early Fusion (data augmentation) & RGB+SAR & 50.00 \\ 
 \hline
\end{tabular}
\end{center}
\end{table}

\vspace{-2em}

\begin{table}[H]
\caption{\label{table3} Height estimation }
\begin{center}
\begin{tabular}{ |c c|c| } 
\hline
  Type & Modality & Score \\
 \hline
 Optical & RGB & 23.88 \\ 
 Optical (optimized) & RGB & 26.31 \\ 
 Early Fusion & RGB+SAR & 30.10 \\ 
 Intermediate Fusion & RGB+SAR & 26.67 \\ 
 Late Fusion & RGB+SAR & 30.20 \\ 
 Early Fusion-Skip Connection & RGB+SAR & 30.65 \\ 
 \hline
\end{tabular}
\end{center}
\end{table}


The best results for height estimation were obtained using Early fusion with skip connections. 
A score of 0.306 indicates that the model explains only 30.6\% of the variance in the dependent variable, which is relatively low. The same variant was able to achieve the lowest Root Mean Square Error (RMSE) of 12.7 on the training set.
The main deficit was under-prediction for larger-height objects such as buildings and over-prediction for lower-height objects like grass. 
Nevertheless, in some cases, a lower R2 score may still be acceptable if the model is providing useful insights or performing well enough for the specific application.


\section{Conclusion}
Although the initial goal of this contest was to solve the problem of estimating the height of buildings extracted in parallel. Using the conventional approach helped to understand the granularity of the problem better. The various experiments made were very promising and with further experiments, the predictability of individual models is sure to improve. A possibility to be explored for better footprints is to create branched auto-encoders that can multitask the learning process while  sharing the weights at an intermediate layer or at a later step within the autoencoder. 
Height estimation models are not up to the mark. Though the state of the art is not exceptionally high, achieves 75\% score on the validation set, yet accuracy score of the best of our fusion approach is not close enough.
There is a gap that is yet to be explored, out of the constraints of the competition, that could be a possibility of improving results. That is 
to see if the results of the building extraction, which seems to improve faster than the height estimation models,  can be fed as input into the height estimation architecture, perhaps further improving the predictive performance of the heights of the buildings once segmented.

\subsection*{\centering\small ACKNOWLEDGEMENT}

\small The authors highly acknowledge Dr. Charlotte Pelletier for her valuable review and guidance. The talks with Prof. Dr. Sébastien Lefèvre, Dr Jean-Christophe Burnel and Dr Frédéric Raimbault in UBS were also helpful in building upon the idea.
The authors would like to thank the IEEE GRSS Image Analysis and Data Fusion Technical Committee, Aerospace Information Research Institute, Chinese Academy of Sciences, Universität der Bundeswehr München, and GEOVIS Earth Technology Co., Ltd. for organizing the Data Fusion Contest
The data used for this research was provided by [REF. NO.] 2023 IEEE GRSS Data Fusion Contest. Online: www.grss-ieee.org/technical-committees/image-analysis-and-data-fusion/
The developed code and output are made availble thorugh https://github.com/SaadAhmedJamal/IEEE\_DFC2023.
This Reserach was support by Erasmus+ Program of the European Union. 

\bibliographystyle{IEEEbib}
\small\bibliography{refs}

\end{document}